\documentclass[letterpaper,10pt,journal,twoside]{IEEEtran}
\usepackage{amsmath}
\usepackage{graphicx}
\usepackage{overpic}
\usepackage{xcolor}
\usepackage{booktabs}
\usepackage{multirow}
\usepackage{tabularx}
\usepackage{array}
\usepackage{cite}
\usepackage[colorlinks=true,linkcolor=blue,citecolor=blue,urlcolor=blue,filecolor=blue]{hyperref}

\title{T-Tac: A Soft and Thin Active Thermal Tactile Sensor for Robotic In-Grasp Material Identification}

\author{Jiankun~Li,~\IEEEmembership{Student Member,~IEEE,}
		Yu~Feng,~\IEEEmembership{Member,~IEEE,}
        Yehui~Liu,~\IEEEmembership{Student Member,~IEEE,}
        Meng~Chen,
        Rui~Luo,
        Guanglie~Zhang,~\IEEEmembership{Senior Member,~IEEE,}
        Jianshu~Zhou,~\IEEEmembership{Member,~IEEE,}
        and~Wen~Jung~Li,~\IEEEmembership{Fellow,~IEEE}%
\thanks{This work was supported by the Hong Kong Research Grants Council (Project Nos. 11222126, 11216120, 11207222, T42-717/20-R, G-CityU105/21, G-CityU104/19, F-CityU104/19, and CRS\_CityU103/25), the Guangdong and Hong Kong Universities 1+1+1 Joint Research Collaboration Scheme, and the InnoHK Project at the Hong Kong Centre for Cerebro-cardiovascular Health Engineering (COCHE). \textit{(Corresponding authors: Yu Feng, Wen Jung Li.)}}%
\thanks{Jiankun Li, Yehui Liu, Meng Chen, Guanglie Zhang, Wen Jung Li are with the Mechanical Engineering Department, City University of Hong Kong, Kowloon, Hong Kong SAR, and CAS-CityU Joint Laboratory for Robotic Research, City University of Hong Kong Shenzhen Research Institute, Shenzhen, China (wenjli@cityu.edu.hk).}%
\thanks{Yu Feng and Jianshu Zhou are with the Department of Mechanical Engineering, National University of Singapore, Singapore (e-mail: yufeng.me@nus.edu.sg).}
\thanks{Rui Luo is with the Department of Systems Engineering, City University of Hong Kong, Kowloon, Hong Kong SAR.}%

}

\begin{document}
\bstctlcite{ttac-reference-format}
\markboth{}%
{Li \MakeLowercase{\textit{et al.}}: A Soft, Thin, and Active Thermal Tactile Sensor}
\maketitle

\begin{abstract}
Accurate material identification is important for robotic manipulation, yet vision-based methods face challenges when materials have similar appearances. Herein, we report T-Tac, a soft, thin, active thermal tactile sensor that identifies materials through their transient thermal responses. T-Tac integrates a polyimide film heater with an ionic conductive film temperature sensor, providing controlled thermal excitation and resistance-based temperature measurement in a film assembly suitable for mounting on robotic grippers. Single-sided transient plane source modeling and finite element simulations relate the thermal response to material thermal properties, while calibration and repeated measurements characterize the heating and sensing performance. Loading experiments at room and elevated temperatures further characterize the sensor response under an applied force. A hybrid classification framework concatenates extracted thermal descriptors with multiscale convolutional neural network features and uses an ExtraTrees classifier for material prediction. Experiments on 12 materials achieve a mean classification accuracy of 92.69\% using a 3 s heating pulse and a 10 s observation window. These results support T-Tac as a compact approach to integrating active thermal material sensing into robotic end effectors. This work shed a light on the new approach for robotic material identification.
\end{abstract}


\section{Introduction}
\IEEEPARstart{I}{n} recent years, the rapid advancement of large language models has significantly enhanced the intelligence of robots in reasoning, planning, and decision-making \cite{kannanSMARTLLMSmartMultiAgent2024}. However, high-level intelligence alone is insufficient to support robotic applications in complex real-world environments. To achieve natural interaction with the environment, robots must not only ``think'' but also ``perceive.'' Among various sensory modalities, vision, audition, and touch are the most essential \cite{xueProgressProspectsMultimodal2020}. In particular, tactile sensing serves as the fundamental channel for physical interaction between robots and their surroundings, providing information on force, vibration, and temperature \cite{ZhouTacSuitWearableLargeArea2024}, and thereby playing a central role in enabling dexterous manipulation and environmental adaptability \cite{kappassovTactileSensingDexterous2015a}.

\begin{figure}[t]
	\centering
	\includegraphics[width=0.90\columnwidth]{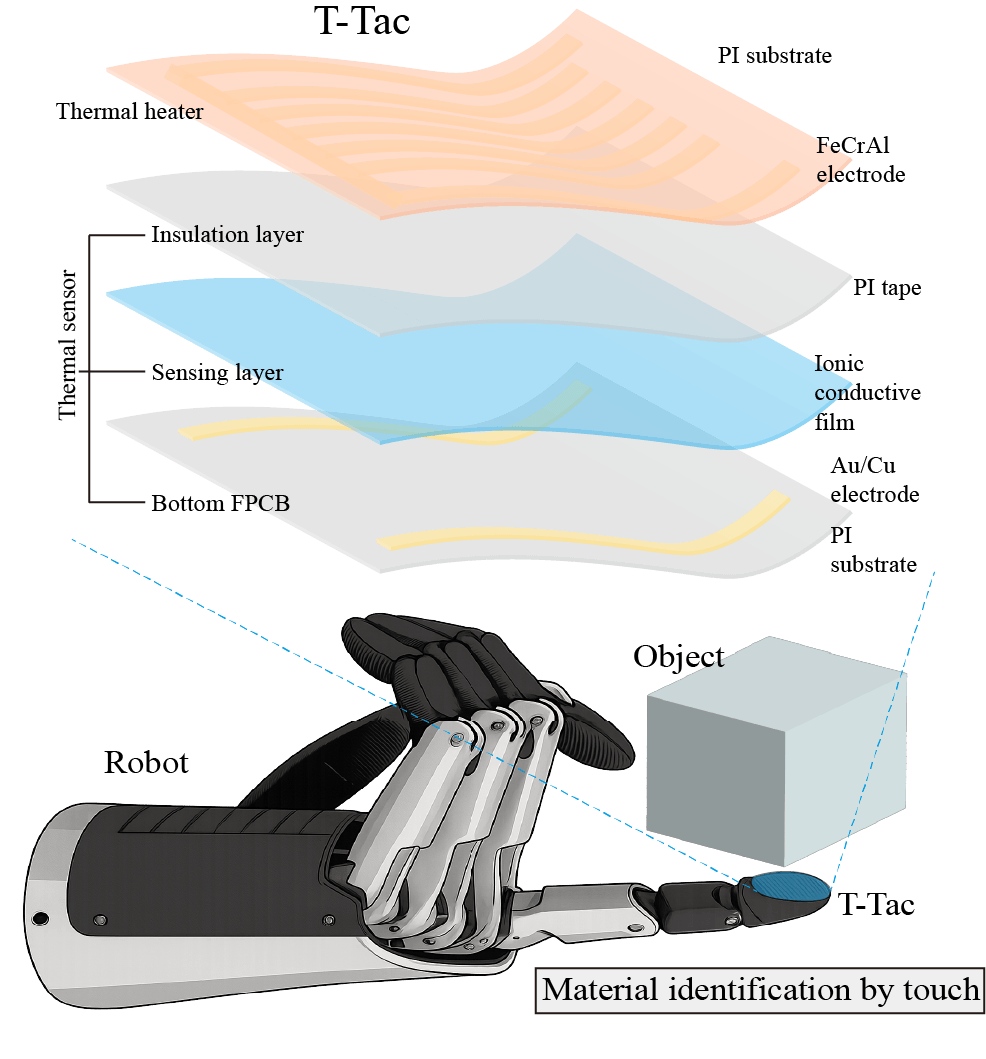}
	\caption{Device design of the soft thermal tactile sensor (T-Tac), which actively generates and senses transient thermal signals to enable rapid material identification by touch.}
	\label{Device Design}
\end{figure}

Benefiting from advances in materials science and nanotechnology, a wide variety of tactile sensors have been developed, including those based on piezoresistive \cite{fonsecaFlexiblePiezoresistiveSelfCapacitive2023,canavesePiezoresistiveFlexibleComposite2014a}, piezoelectric \cite{ShinArtificialTactileSensor2020}, capacitive  \cite{maiolinoFlexibleRobustLarge2013,liuMaterialRecognitionUsing2024}, electromagnetic \cite{xieDevelopmentTriaxisElectromagnetic2022}, magnetic \cite{FengBioinspiredSoftMagnetociliary2026,LiuInverseDesignMagnetic2026}, and optical imaging \cite{FengSoftMechanoluminescentSkin2026}. These sensors endow robots with diverse tactile capabilities, such as force perception, vibration detection, and surface texture recognition. However, in scenarios such as safe grasping, material classification, and human--robot interaction \cite{liuRoboticMaterialPerception2019,takamukuRobustMaterialDiscrimination2008}, the ability to identify materials is often required. Although vision-based algorithms can recognize the majority of materials, their accuracy drops significantly when facing objects with similar appearances. Even sliding or friction-based tactile sensors, which can capture subtle differences in surface texture \cite{WangUltrasensitiveTorsionTactile2026}, remain limited in distinguishing materials with highly similar surface features.

Humans rely mainly on thermal touch when distinguishing materials with similar appearances \cite{kappersThermalPerceptionThermal2019,jonesMaterialDiscriminationThermal2003,adelsonSeeingStuffPerception2001}. When thermoregulated human skin comes into contact with different objects, cutaneous thermoreceptors sense transient temperature changes and use this information to infer the material category. Temperature sensing has also been used for contact detection in continuum robots \cite{YangTemperatureBasedContact2026}. Inspired by the human mechanism of actively perceiving materials through thermal touch, we propose a soft thermal tactile sensor (T-Tac). The sensor consists of a polyimide (PI) film-based thermal heater and an ionic conductive film-based thermal sensor. T-Tac is lightweight and ultrathin, making it easy to integrate onto robotic end-effectors. In addition to providing controlled thermal stimulation, T-Tac offers highly sensitive temperature measurements. By applying pulsed thermal excitation via the thermal heater and capturing the transient thermal response through the thermal sensor, T-Tac enables robots to identify different materials. Furthermore, leveraging the single-sided transient plane source method combined with thermal-feature and CNN-feature fusion using an ExtraTrees classifier, we established a classification model for 12 materials, enabling reliable material identification through thermal touch. This work opens new avenues for the design of thermal tactile sensors and provides a novel strategy for imparting human-like tactile perception to robots.

The remainder of this paper is structured as follows. Section \ref{Design of T-Tac} introduces the overall design concept of T-Tac, and also covers the fabrication of the ionic conductive film together with the hardware architecture. The operating mechanisms of the thermal heater and the thermal sensor are explained in Section \ref{Modeling of T-Tac}, where theoretical modeling and simulations based on the single-sided transient plane source method are also carried out. Experimental calibration results of both the thermal heater and the thermal sensor are presented in Section \ref{Performance of T-Tac}. Section~\ref{Verification Experiments} reports material-classification experiments and illustrative gripper-mounted measurements. Then, the comparison with related works for material identification is presented in Section~\ref{Comparison with Related Works}. Finally, Section \ref{Conclusion} summarizes this work.

\section{Design of T-Tac}
\label{Design of T-Tac}
\subsection{Device Design}
Material-dependent transient heat transfer provides a basis for thermal tactile discrimination \cite{wuQuantitativeEvaluationNatural2023}. T-Tac combines a heater for thermal excitation with a temperature sensor for recording the contact response. As shown in Fig.~\ref{Device Design}, the heater comprises a PI substrate and a FeCrAl electrode. The temperature sensor uses a sandwich structure comprising PI tape, an ionic conductive film, and a flexible printed circuit board (FPCB). This film-based assembly supports attachment to robotic end effectors.

\subsection{Device Fabrication}
\begin{figure}[htbp]
	\centering
	\includegraphics[width=0.855\columnwidth]{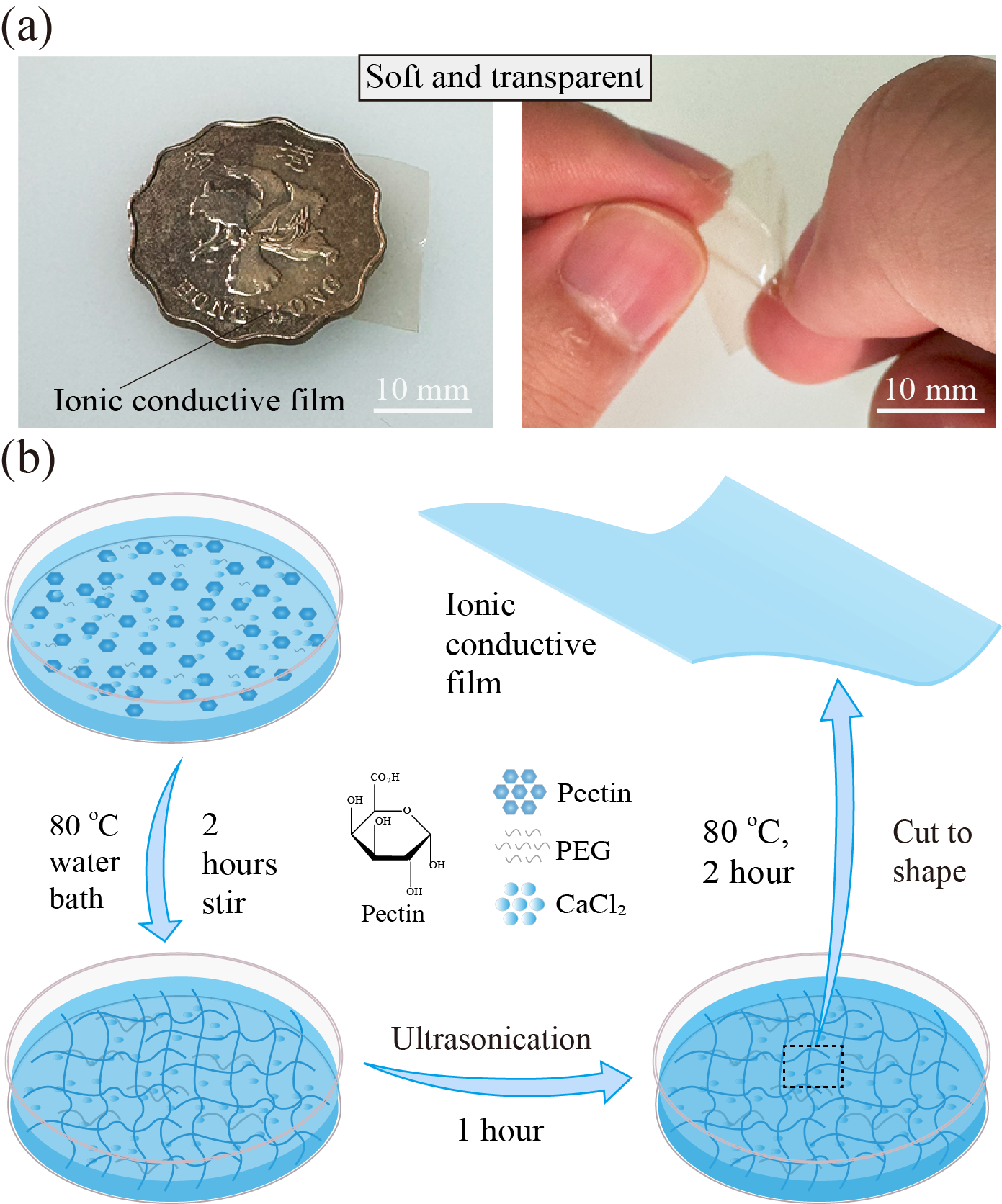}
	\caption{Device fabrication of T-Tac. (a) Optical photographs of the ionic conductive film. (b) Schematic fabrication process of the ionic conductive film.}
	\label{Device Fabrication}
\end{figure}

The thermal heater with dimensions of $20 \times 20 \times 0.15 \ \text{mm}$ and resistance of $10 \ \Omega$ was custom-provided by a supplier, while the thermal sensor adopted a simple sandwich structure consisting of a PI tape as the top insulating layer, an ionic conductive film as the sensing layer, and an FPCB as the bottom electrode. The PI tape and FPCB were supplied by JLC Group, so only the ionic conductive film needed to be fabricated. As shown in Fig.~\ref{Device Fabrication}(a), the ionic conductive film is a soft and transparent thin film, and its preparation followed a simple one-pot method \cite{FengIntrinsicForceTemperature2026}. As shown in Fig.~\ref{Device Fabrication}(b), pectin, polyethylene glycol (PEG), and calcium chloride were dissolved in deionized water, heated in a water bath at $80 \ ^\circ \text{C}$, and stirred for $2 \ \text{h}$ until complete dissolution and reaction were achieved, followed by ultrasonic treatment for $1 \ \text{h}$ to remove air bubbles, and finally dried at $80 \ ^\circ \text{C}$ for $2 \ \text{h}$ to obtain the ionic conductive film. The fabrication of T-Tac was completed by tightly attaching the thermal heater onto the surface of the encapsulated thermal sensor.

\subsection{Hardware Design and System Integration}
\begin{figure}[htbp]
	\centering
	\includegraphics[width=0.90\columnwidth]{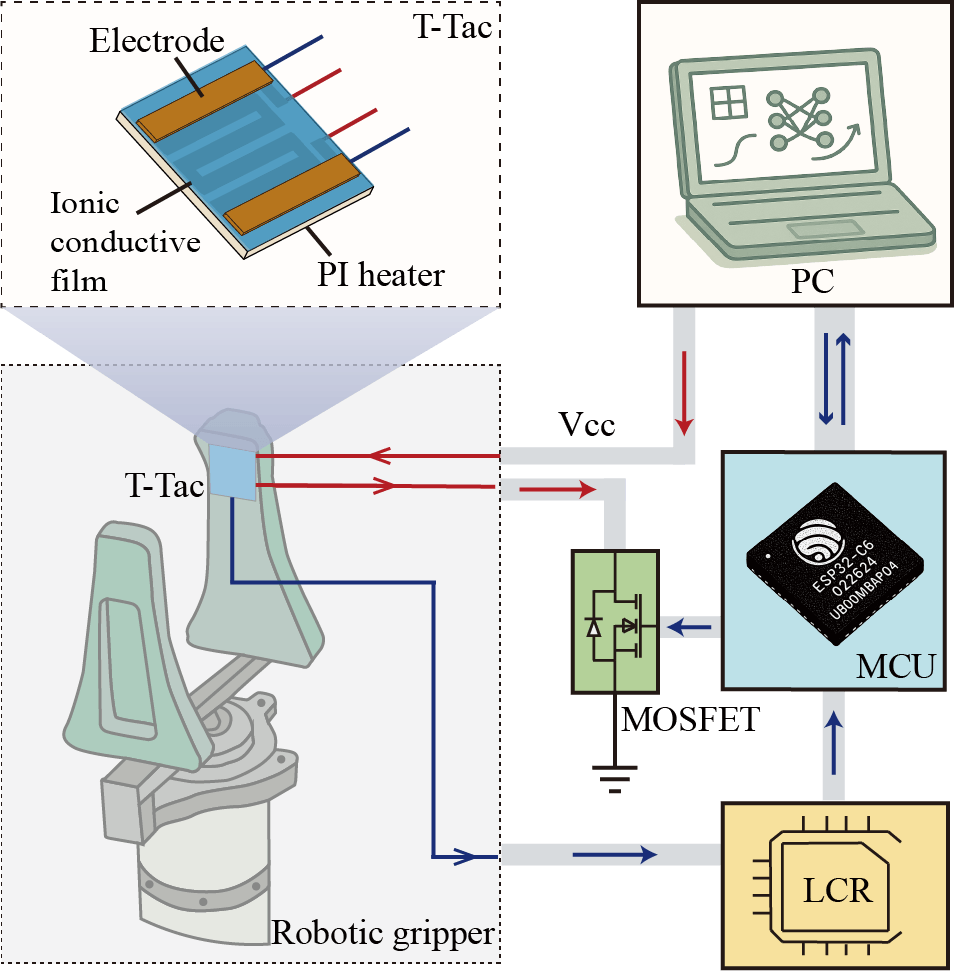}
    \caption{Hardware design and system integration of T-Tac, showing PWM-controlled heating, resistance acquisition, and data transfer to the host computer.}
	\label{Hardware Design}
\end{figure}

The hardware architecture of T-Tac consists of a sensing module, a data acquisition module, and a data processing module, as shown in Fig.~\ref{Hardware Design}. The core control unit of the system is an ESP32-C6 microcontroller (MCU). Resistance measurement is performed by an M180 LCR bridge module, and the power switching device is an AO3400A N-channel MOSFET. One end of the PI heating film is connected to a $5 \ \text{V}$ power supply from a USB interface, and the other end is connected to the control circuit through the MOSFET. The MCU regulates the duty cycle of the heating film by switching the MOSFET on and off, thereby indirectly adjusting the heating temperature. As the thermal heater causes a temperature change, the resistance of the thermal sensor responds accordingly. The LCR module collects the resistance value in real time and transmits the data to the MCU, which then uploads the data to a computer for further processing and analysis.

\section{Modeling of T-Tac}
\label{Modeling of T-Tac}
\subsection{Principle of Thermal Heater}
\begin{figure}[htbp]
	\centering
	\includegraphics[width=\columnwidth]{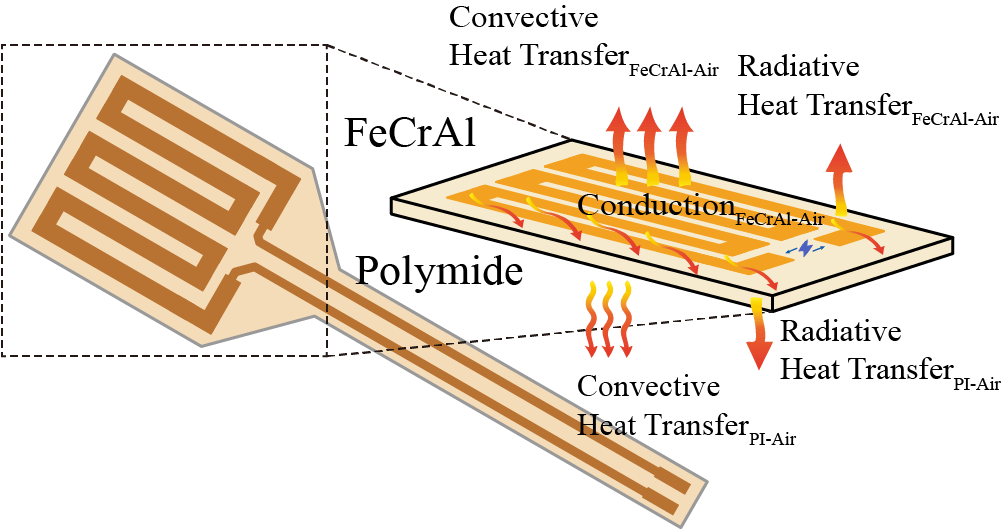}
	\caption{Principle of the thermal heater, a thin PI-based heating film that generates controlled heat through Joule heating while exchanging energy via conduction, convection, and radiation.}
	\label{Principle of Thermal Heater}
\end{figure}

The thermal heater generates Joule heat in a metallic circuit printed on a PI film (Fig.~\ref{Principle of Thermal Heater}) \cite{huangHighlyFlexibleTransparent2015}. Heat is conducted through the film and dissipated to the surroundings. At steady state, neglecting radiation at low operating temperatures, the balance between electrical input and conductive and convective heat transfer is expressed as

{\begin{equation}
R I^{2} = \lambda_{m} S_{m} \frac{\Delta T}{\Delta x}
+ h_{a} S_{a} \Delta T.
\end{equation}
}
Here, $R$ and $I$ are the heater resistance and current, $\lambda_m$ and $S_m$ are the thermal conductivity and effective conduction area of the film, and $h_a$ and $S_a$ are the convective heat transfer coefficient and area. The temperature difference $\Delta T$ over the conduction length $\Delta x$ defines the thermal gradient. Thus, the heater temperature can be adjusted through its input power.

\subsection{Principle of Thermal Sensing and Transient Thermal Response}
\begin{figure}[htbp]
    \centering
    \resizebox{0.90\columnwidth}{!}{%
\begin{overpic}[width=\columnwidth,abs,unit=0.01\columnwidth]{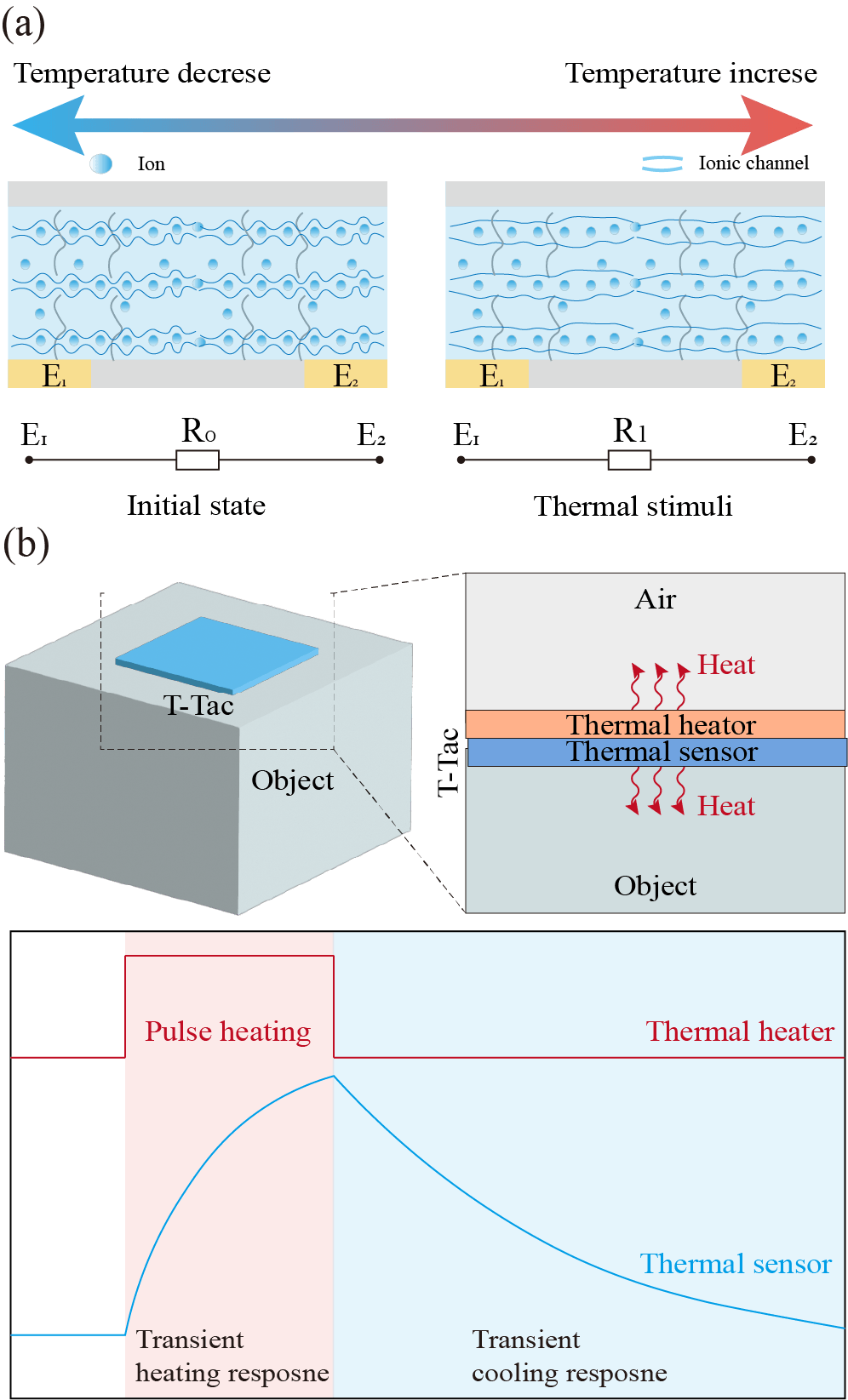}
        \put(1.0040,154.5181){\makebox(0,0)[bl]{\begingroup\setlength{\fboxsep}{0pt}\colorbox{white}{\parbox[c][4.0161\unitlength][c]{32.1285\unitlength}{\centering\fontsize{8.10}{8.60}\selectfont Temperature decrease}}\endgroup}}
        \put(66.0643,154.5181){\makebox(0,0)[bl]{\begingroup\setlength{\fboxsep}{0pt}\colorbox{white}{\parbox[c][4.0161\unitlength][c]{33.6345\unitlength}{\centering\fontsize{8.10}{8.60}\selectfont Temperature increase}}\endgroup}}
        \put(65.7631,78.2129){\makebox(0,0)[bl]{\begingroup\setlength{\fboxsep}{0pt}\colorbox[RGB]{255,177,138}{\parbox[c][3.0120\unitlength][c]{28.6145\unitlength}{\centering\fontsize{8.10}{8.60}\selectfont Thermal heater}}\endgroup}}
    \end{overpic}%
}
    \caption{Principle of thermal sensing and transient thermal response. (a) Temperature-dependent ion migration and resistance variation in the ionic conductive film. (b) Single-sided thermal contact and the heating and cooling response produced by pulsed excitation.}
    \label{Thermal Sensing and Transient Response}
    \label{Principle of Thermal Sensor}
    \label{Mechanism of Transient Thermal Response}
\end{figure}

T-Tac converts the transient temperature at the contact interface into a resistance signal through an ionic conductive film. As shown in Fig.~\ref{Thermal Sensing and Transient Response}(a), calcium ions migrate through conductive channels in the crosslinked pectin network \cite{digiacomoBiomimeticTemperaturesensingLayer2017}. Increasing temperature enhances ion migration and lowers the resistance between electrodes $E_1$ and $E_2$ from $R_0$ to $R_1$; cooling raises the resistance toward $R_0$. This negative temperature coefficient allows the film to capture both heating and cooling responses.

As shown in Fig.~\ref{Thermal Sensing and Transient Response}(b), a brief heating pulse raises the interface temperature, which decays after the pulse as heat dissipates into the sample and surroundings. The resulting temperature history depends on the material's thermal properties and is measured through the film resistance. This process follows the transient plane source (TPS) principle \cite{logTransientPlaneSource1995}, with T-Tac contacting one surface of the sample in the single-sided configuration \cite{aiSinglesidedMethodBased2017}. For a short pulse, heat transfer is approximated by conduction into semi-infinite media with constant properties on each side. Thermal diffusivity is $\kappa=\lambda/(\rho c)$, where $\lambda$, $\rho$, and $c$ denote thermal conductivity, density, and specific heat capacity, respectively.

Equating the interface temperature responses on the two sides and applying the Laplace transform gives the power delivered to the sample:
{\begin{equation}
P_s=P_0\left[1+\frac{\lambda_b}{\lambda}
\left(\frac{\kappa}{\kappa_b}\right)^{1/2}\right]^{-1}.
\end{equation}
}
Here, $P_0$ is the total heating power, and the subscript $b$ denotes the background medium. The power partition is constant under these assumptions. Substituting $P_s$ into the TPS temperature response yields
{\begin{equation}
\Delta\overline{T}(\xi)=
\frac{2P_0 D(\xi)}{\pi^{3/2}a\lambda
\left[1+(\lambda_b/\lambda)(\kappa/\kappa_b)^{1/2}\right]}.
\end{equation}
}
Here, $a$ is the characteristic size of the planar heat source, $\xi=\sqrt{\kappa t}/a$ is dimensionless time, and $D(\xi)$ is the TPS response function. The variable $t$ denotes elapsed heating time. The idealized relation links the contact response to material thermal properties under the stated semi-infinite assumptions. Finite sensor geometry, heat capacity, and contact thermal resistance affect its use for quantitative property inference. T-Tac uses the measured transient response as input to material classification.

\subsection{Simulation of Transient Thermal Response}
Finite element simulations were performed in COMSOL for 12 materials using the experimental sensor and sample dimensions. T-Tac applied a $2.5\ \text{s}$ heating pulse, and the contact-interface temperature was recorded during heating and cooling. With $V=5\ \text{V}$ and $R=10\ \Omega$, the electrical power and volumetric heat source are

\begin{figure}[htbp]
	\centering
	\includegraphics[width=\columnwidth]{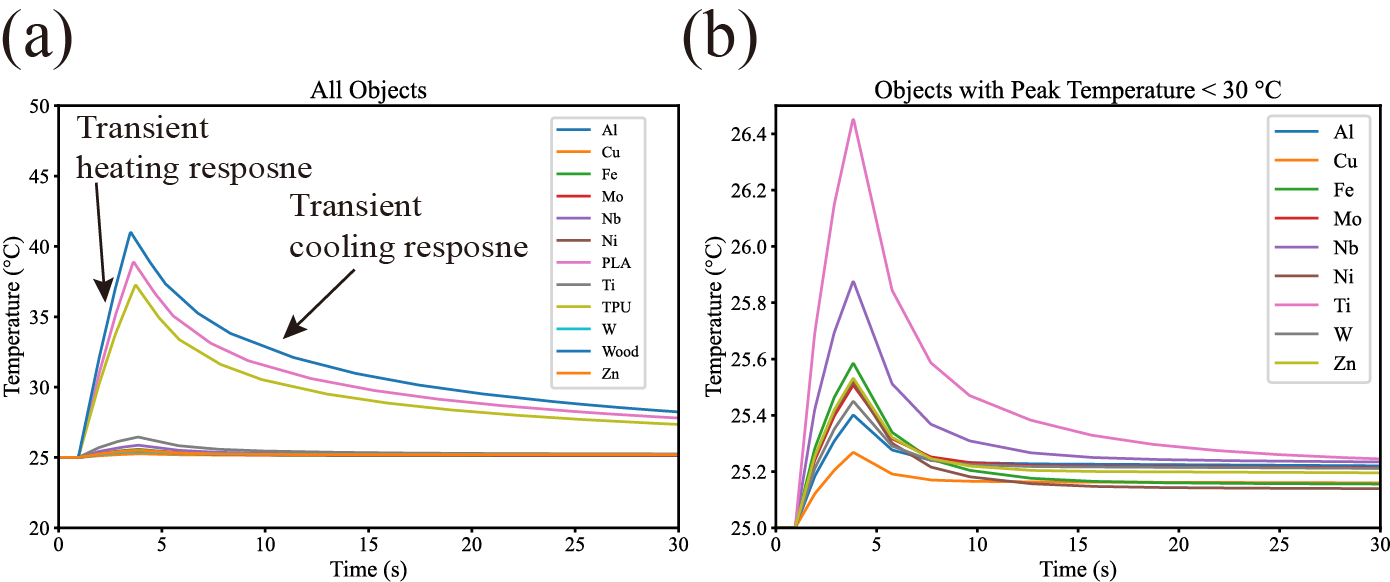}
	\caption{Simulation of the transient thermal response for 12 different materials, showing distinct heating and cooling profiles that reflect their material-dependent thermal properties.}
	\label{Simulation of Transient Thermal Response}
\end{figure}

{\begin{equation}
\begin{aligned}
P &= V^2/R=2.5\ \text{W},\\
Q_{\mathrm{vol}}(t) &=\frac{P}{V_h}
\begin{cases}
1, & 0\leq t\leq 2.5\ \text{s},\\
0, & \text{otherwise},
\end{cases}\\
P/V_h &= 3.125\times10^7\ \mathrm{W\,m^{-3}}.
\end{aligned}
\end{equation}
}
The heated volume is $V_h=(4.0\times10^{-4})(2.0\times10^{-4})\ \mathrm{m^3}$. Both the sensor and samples were initially at $25\ ^\circ\text{C}$, and air-exposed surfaces exchanged heat by natural convection. The material properties are listed in Table~\ref{Thermal properties of tested materials}.

\begin{table}[t]
\centering
\caption{Thermal properties of tested materials \cite{grimvallThermophysicalPropertiesMaterials1999}}
\label{Thermal properties of tested materials}
\footnotesize
\setlength{\tabcolsep}{2.5pt}
\renewcommand{\arraystretch}{1.15}
\begin{tabular*}{\columnwidth}{@{}l@{\extracolsep{\fill}}cccc@{}}
\hline\hline
\textbf{Material} & \textbf{Symbol} & 
\shortstack{$\boldsymbol{k}$\\(W·m$^{-1}$·K$^{-1}$)} & 
\shortstack{$\boldsymbol{C_p}$\\(J·kg$^{-1}$·K$^{-1}$)} & 
\shortstack{$\boldsymbol{\rho}$\\(kg·m$^{-3}$)} 
\\
\hline
Aluminum                   & Al   & 205   & 900  & 2700  \\
Copper                     & Cu   & 385   & 385  & 8960  \\
Iron                       & Fe   & 80    & 450  & 7870  \\
Molybdenum                 & Mo   & 138   & 251  & 10200 \\
Niobium                    & Nb   & 54    & 265  & 8570  \\
Nickel                     & Ni   & 91    & 444  & 8900  \\
Polylactic Acid            & PLA  & 0.13  & 1800 & 1240  \\
Titanium                   & Ti   & 22    & 522  & 4500  \\
\begin{tabular}[c]{@{}l@{}}Thermoplastic\\Polyurethane\end{tabular} & TPU  & 0.20  & 1500 & 1200  \\
Tungsten                   & W    & 174   & 134  & 19300 \\
Wood                       & Wood   & 0.12  & 1700 & 600 \\
Zinc                       & Zn   & 116   & 388  & 7140  \\
\hline\hline
\end{tabular*}
\end{table}

As shown in Fig.~\ref{Simulation of Transient Thermal Response}(a), the temperature rises during heating and decays after the pulse. The cooling response is described by
{\begin{equation}
T(t)=T_{\infty}+A e^{-t/\tau},\qquad A=T_{\max}-T_{\infty}.
\end{equation}
}
Here, $t$ is measured from the start of cooling, $T_{\infty}$ is the asymptotic temperature, and $\tau$ is the time required for the excess temperature to fall to $1/e$ of its initial value. The extracted parameters are summarized in Table~\ref{Thermal response parameters of tested materials}.

\begin{table}[t]
\centering
\caption{Thermal response parameters of tested materials}
\label{Thermal response parameters of tested materials}
\footnotesize
\setlength{\tabcolsep}{2.5pt}
\renewcommand{\arraystretch}{1.15}
\begin{tabular*}{\columnwidth}{@{}l@{\extracolsep{\fill}}ccccc@{}}
\hline\hline
\multirow{2}*{\centering\textbf{Object}} &
\multirow{2}*{\shortstack{\textbf{Peak Temp}\\\textbf{(°C)}}} &
\multirow{2}*{\shortstack{\textbf{Heating Rate}\\\textbf{(°C/s)}}} &
\multicolumn{3}{c}{$T(t)=T_\infty+A\exp(-t/\tau)$} \\
& & & \shortstack{\textbf{Time constant}\\$\boldsymbol{\tau}$ \textbf{(s)}} & \shortstack{$\boldsymbol{T_\infty}$\\\textbf{(°C)}} & \shortstack{\textbf{A}\\\textbf{(°C)}} \\
\hline
Al   & 25.40 & 0.12 & 1.70 & 25.22 & 1.83 \\
Cu   & 25.27 & 0.08 & 1.61 & 25.16 & 1.26 \\
Fe   & 25.58 & 0.18 & 2.47 & 25.16 & 2.06 \\
Mo   & 25.51 & 0.15 & 1.90 & 25.22 & 2.32 \\
Nb   & 25.87 & 0.27 & 2.49 & 25.24 & 3.02 \\
Ni   & 25.52 & 0.16 & 2.45 & 25.14 & 1.84 \\
PLA  & 38.87 & 3.90 & 6.24 & 28.01 & 17.94 \\
Ti   & 26.45 & 0.45 & 3.24 & 25.28 & 3.75 \\
TPU  & 37.24 & 3.40 & 5.81 & 27.57 & 16.87 \\
W    & 25.45 & 0.14 & 1.80 & 25.21 & 2.12 \\
Wood & 40.98 & 4.63 & 6.96 & 28.35 & 19.35 \\
Zn   & 25.53 & 0.16 & 2.04 & 25.20 & 2.30 \\
\hline\hline
\end{tabular*}
\end{table}

Wood, PLA, and TPU exhibit higher peak temperatures than the nine metals because their lower thermal conductivities promote heat accumulation at the interface. The metals show smaller temperature rises and faster cooling (Fig.~\ref{Simulation of Transient Thermal Response}(b)). Among them, titanium has the highest peak owing to its relatively low conductivity and diffusivity, while copper has the lowest peak owing to rapid heat transfer into the bulk. These differences provide thermal features for material classification.

\section{Performance of T-Tac}
\label{Performance of T-Tac}
\subsection{Performance of Thermal Heater}
\begin{figure}[htbp]
	\centering
	\includegraphics[width=0.90\columnwidth]{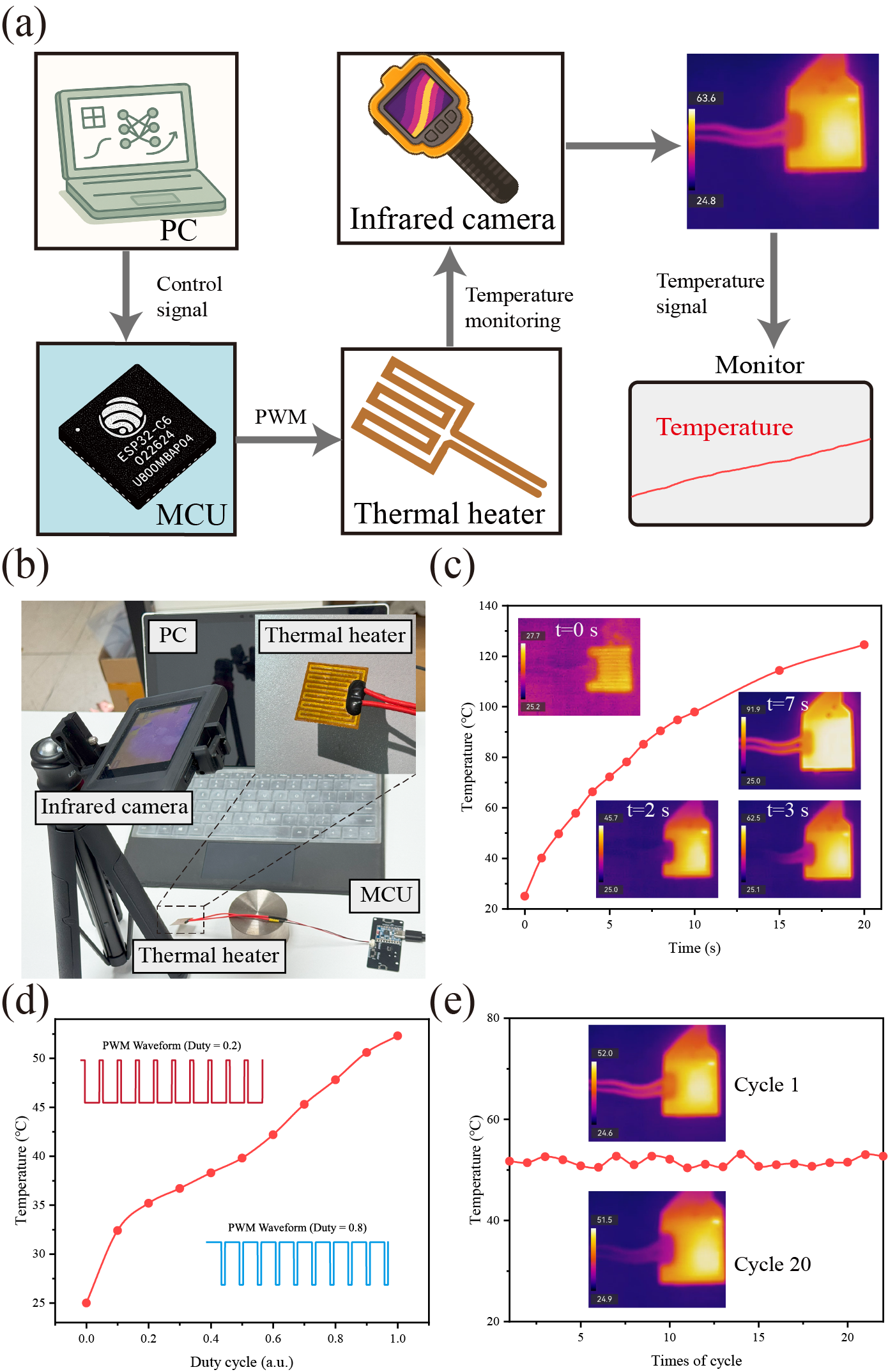}
	\caption{Calibration and performance of the thermal heater. (a) Schematic of the calibration setup. (b) Experimental platform. (c) Temperature variation with heating duration. (d) Maximum temperature at different duty cycles. (e) Maximum temperature during repeated heating.}
	\label{Calibration and Performance of Thermal Heater}
\end{figure}

The thermal heater was calibrated using the platform shown in Fig.~\ref{Calibration and Performance of Thermal Heater}(a) and (b). An MCU controlled the heater, while an infrared camera monitored its temperature in real time.

At $5\ \text{V}$, the heater temperature increases nonlinearly with heating duration as heat dissipation increases (Fig.~\ref{Calibration and Performance of Thermal Heater}(c)). For a fixed duration of $3\ \text{s}$, the maximum temperature rises approximately linearly with PWM duty cycle (Fig.~\ref{Calibration and Performance of Thermal Heater}(d)), enabling control of the thermal excitation. Repeated heating at $5\ \text{V}$ for $3\ \text{s}$ produces similar peak temperatures with small fluctuations (Fig.~\ref{Calibration and Performance of Thermal Heater}(e)), demonstrating repeatable thermal stimulation.

\subsection{Performance of Thermal Sensor}
\begin{figure}[htbp]
	\centering
	\includegraphics[width=0.90\columnwidth]{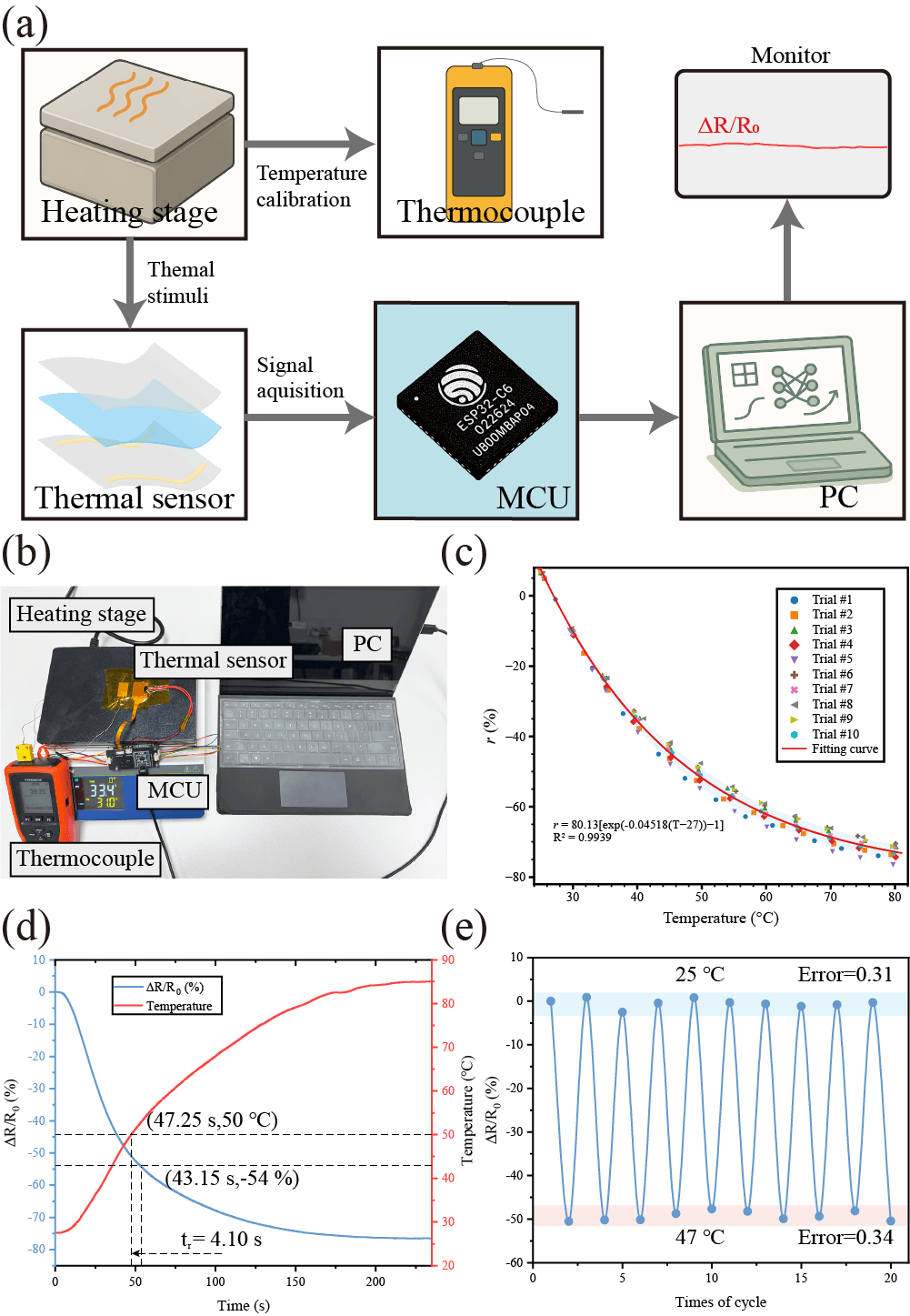}
	\caption{Calibration and performance of thermal sensor. (a) Schematic illustration of calibration. (b) Experimental platform of the thermal sensor. (c) Sensing calibration curves of the thermal sensor. (d) Dynamic performance of the thermal sensor. (e) Repeatability performance of the thermal sensor.}
	\label{Calibration and Performance of Thermal Sensor}
\end{figure}

The thermal sensor was calibrated on a heating stage using a thermocouple as the temperature reference (Fig.~\ref{Calibration and Performance of Thermal Sensor}(a) and (b)). The stage temperature was increased in $2\ ^\circ\text{C}$ increments, and the normalized resistance change was recorded. The reference resistance $R_0$ was measured at room temperature. Consistent with Fig.~\ref{Calibration and Performance of Thermal Sensor}(c), the calibration response is expressed as $r=100(R-R_0)/R_0$, in percent. Ten calibration measurements show a decreasing exponential response; Table~\ref{Sensing parameters of the thermal sensor} summarizes the fitted relation, where $x$ is temperature in degrees Celsius.

\begin{figure*}[!t]
    \centering
    \includegraphics[width=\textwidth]{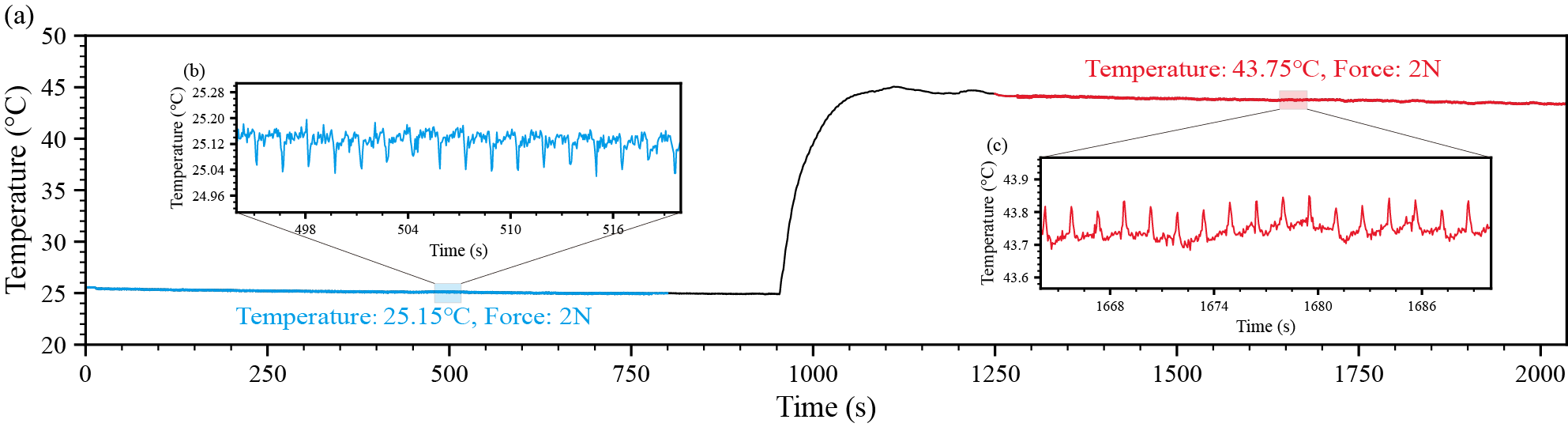}
    \caption{Temperature response of the thermal sensor under a force of $2\ \text{N}$. (a) Temperature recorded during the room-temperature and heated stages. (b) Enlarged response near $25.15\ ^\circ\text{C}$. (c) Enlarged response near $43.75\ ^\circ\text{C}$.}
    \label{Thermal Sensor Repeated Loading}
\end{figure*}

For the dynamic test, T-Tac and the heating stage were initially at $25\ ^\circ\text{C}$. The stage was set to $80\ ^\circ\text{C}$, while the thermocouple measured a surface temperature of $85\ ^\circ\text{C}$ at $235\ \text{s}$ because the stage control point and surface measurement point differed. Using the thermocouple as the reference, the sensor resistance decreases exponentially as the stage heats (Fig.~\ref{Calibration and Performance of Thermal Sensor}(d)), with the response time listed in Table~\ref{Sensing parameters of the thermal sensor}.

Repeatability was evaluated over ten cycles between room temperature and $47\ ^\circ\text{C}$, calibrated by the thermocouple (Fig.~\ref{Calibration and Performance of Thermal Sensor}(e)). The standard errors $\sigma/\sqrt{n}$ are $0.31$ and $0.34$ at $25\ ^\circ\text{C}$ and $47\ ^\circ\text{C}$, respectively, indicating consistent resistance responses across cycles. The temperature response was further examined under a force of $2\ \text{N}$ at room and elevated temperatures (Fig.~\ref{Thermal Sensor Repeated Loading}(a)). The signal remains near $25.15\ ^\circ\text{C}$ before heating and reaches approximately $43.75\ ^\circ\text{C}$ afterward, with recurring fluctuations around the local baseline in the enlarged curves (Fig.~\ref{Thermal Sensor Repeated Loading}(b) and (c)). These measurements characterize temperature sensing under the applied force at both temperature levels.

\begin{table}[t]
\centering
\footnotesize
\setlength{\tabcolsep}{2pt}
\renewcommand{\arraystretch}{1.15}
\caption{Sensing parameters of the thermal sensor}
\label{Sensing parameters of the thermal sensor}
\begin{tabular*}{\columnwidth}{@{}l@{\extracolsep{\fill}}l@{}}
\hline\hline
\textbf{Sensing Parameter} & \textbf{Result} \\
\hline
Range                      & 25--80\,$^\circ$C \\
Fitting equation           & $r=80.13[\exp(-0.04518(x-27))-1]$ \\
Average sensitivity        & $-1.46\%/^\circ$C \\
Coefficient of determination & 0.99 \\
Response time              & 4.10\,s \\
Repeatability error        & 0.34 \\
\hline\hline
\end{tabular*}
\end{table}

\section{Verification Experiments}
\label{Verification Experiments}
\subsection{Experimental Platform and Data Acquisition}
\begin{figure}[htbp]
	\centering
	\includegraphics[width=0.90\columnwidth]{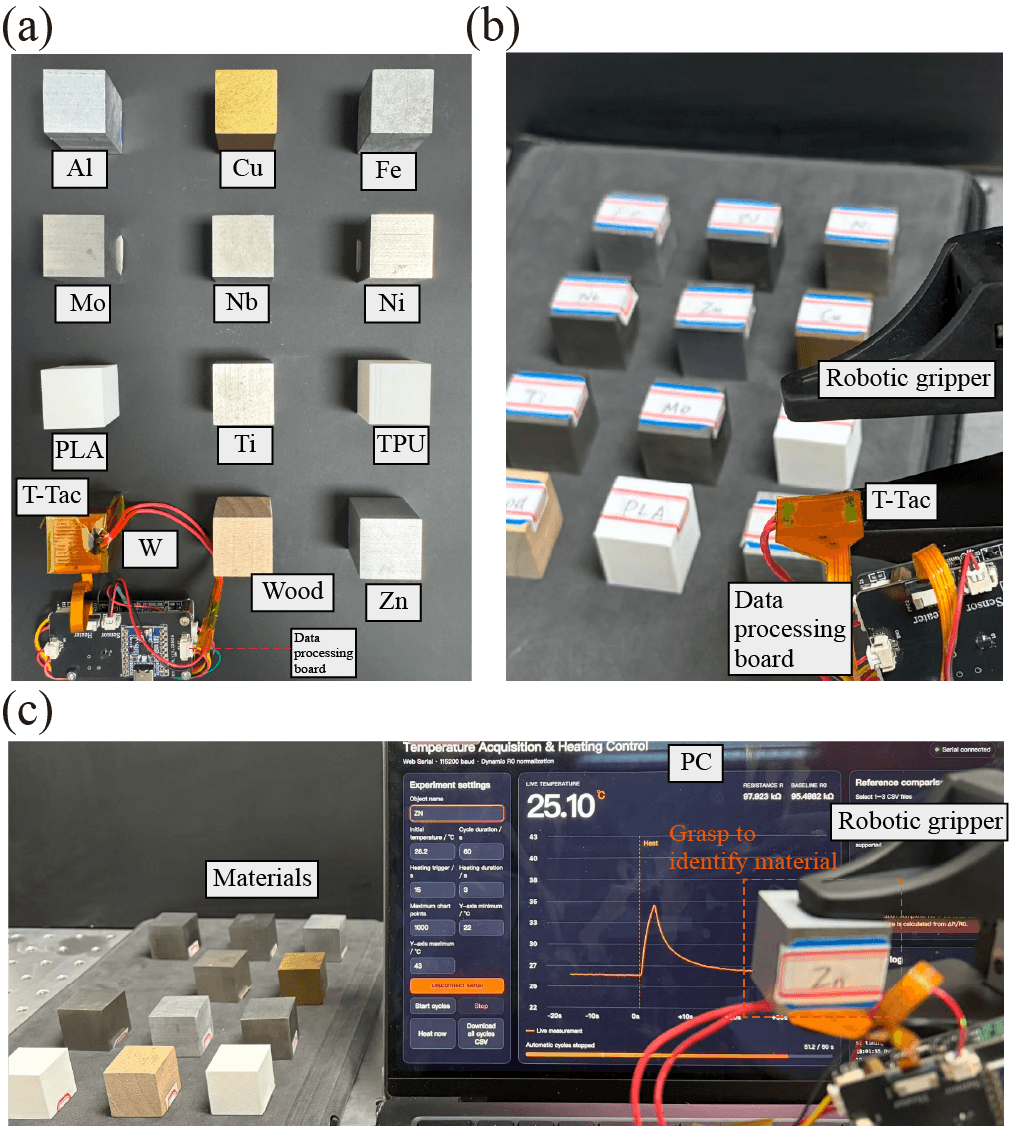}
    \caption{Experimental platform for thermal tactile measurements. (a) Tested materials. (b) T-Tac mounted on a robotic gripper. (c) Measurement setup with the displayed transient thermal response during pulsed heating and cooling.}
	\label{Experimental Platform}
\end{figure}

To validate the effectiveness of material recognition using the single-sided TPS method with T-Tac, experiments were conducted on 12 materials as illustrated in Fig.~\ref{Experimental Platform}(a). In each experiment, T-Tac was attached to the top surface of a small cube, and for every material, 30 independent tests (360 recordings in total) were performed under identical conditions. During the measurement process, T-Tac first applied a controlled heating pulse of $3 \ \text{s}$ at $5 \ \text{V}$ and then naturally cooled to room temperature (RT), while capturing the complete heating and cooling transient response. This controlled thermal excitation ensured that the thermal responses of different materials were obtained under consistent conditions, effectively reducing the influence of environmental disturbances on the experimental results.

The experimental platform for validating the deployment of T-Tac on a robotic end effector is shown in Fig.~\ref{Experimental Platform}(b). In this setup, T-Tac was attached to the inner surface of the robotic gripper using PI tape, and the sensing data were transmitted to a PC through a data processing board, as illustrated in Fig.~\ref{Experimental Platform}(c). During the experiments, the robot grasped an object and employed the single-sided TPS method to acquire its transient thermal response curve, after which material identification was performed using the developed hybrid physics-data-driven algorithm framework.

\subsection{Hybrid Thermal-Feature and Time-Series Classification Framework}
\begin{figure}[!t]
    \centering
    \resizebox{0.90\columnwidth}{!}{%
\begin{overpic}[width=\columnwidth,abs,unit=0.01\columnwidth]{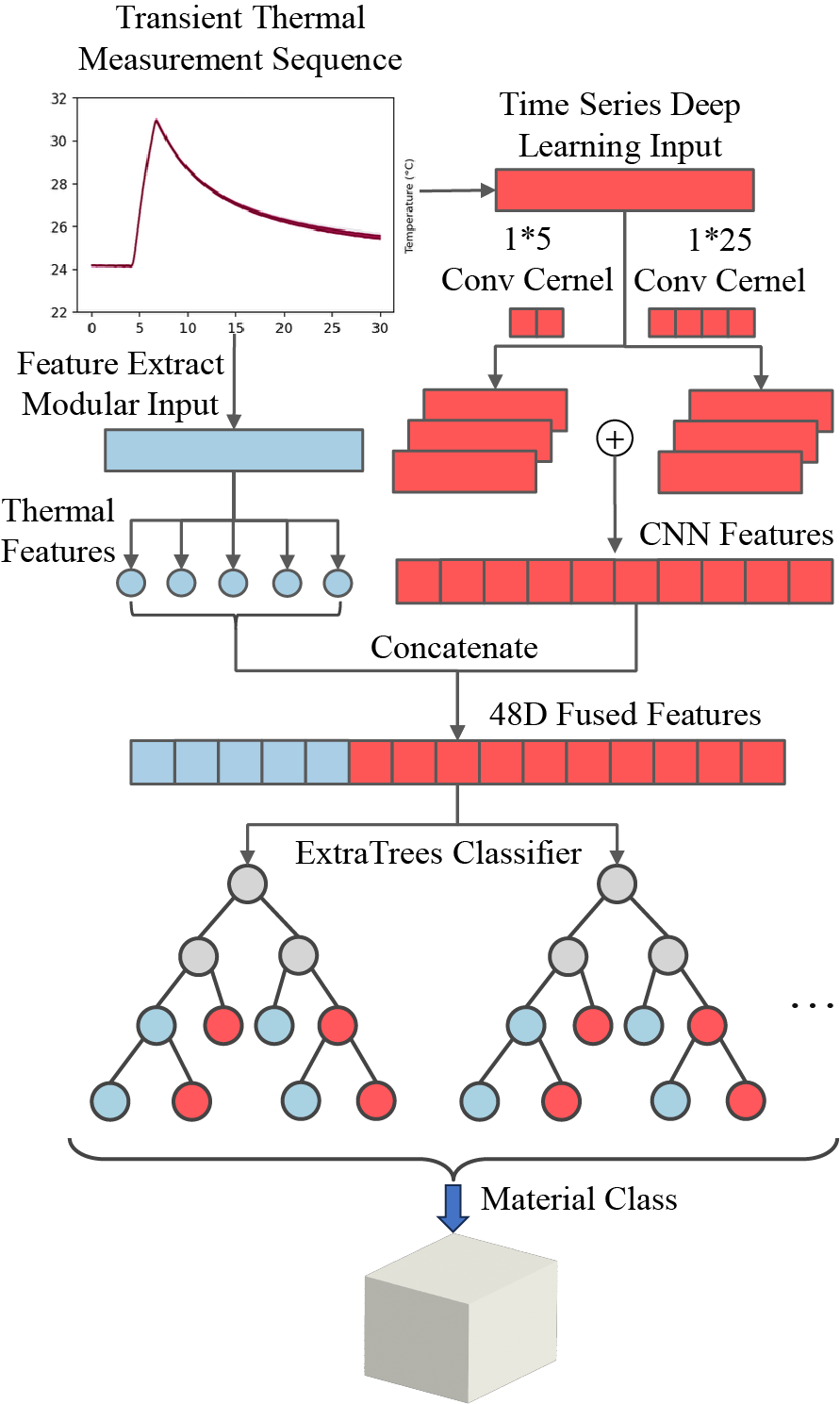}
        \put(51.7396,131.8743){\makebox(0,0)[bl]{\begingroup\setlength{\fboxsep}{0pt}\colorbox{white}{\parbox[c][4.0404\unitlength][c]{21.7733\unitlength}{\centering\fontsize{8.40}{8.90}\selectfont Conv Kernel}}\endgroup}}
        \put(74.7475,131.8743){\makebox(0,0)[bl]{\begingroup\setlength{\fboxsep}{0pt}\colorbox{white}{\parbox[c][4.0404\unitlength][c]{22.5589\unitlength}{\centering\fontsize{8.40}{8.90}\selectfont Conv Kernel}}\endgroup}}
        \put(0.8979,116.2738){\makebox(0,0)[bl]{\begingroup\setlength{\fboxsep}{0pt}\colorbox{white}{\parbox[c][9.4276\unitlength][c]{26.0382\unitlength}{\centering\fontsize{8.40}{8.90}\selectfont Thermal Feature\\Extraction}}\endgroup}}
    \end{overpic}%
}
    \caption{Material classification framework based on thermal and CNN feature fusion. Thermal descriptors and multiscale CNN features are concatenated and fed into an ExtraTrees classifier for material identification.}
    \label{Algorithm Framework}
\end{figure}

Learning-based methods have been explored for tactile reconstruction \cite{DongEfficientTactilePerception2026} and pressure recognition \cite{ChuBilinearFeatureFusion2022}. The proposed classification framework combines extracted thermal descriptors with learned time-series representations to identify materials from their transient thermal responses (Fig.~\ref{Algorithm Framework}). The same measured temperature sequence is supplied to two parallel feature-extraction pathways. In the thermal-feature pathway, the feature-extraction module converts the transient measurement into a vector of thermal descriptors. In the time-series pathway, a convolutional neural network (CNN) uses convolutional kernels of sizes $1\times5$ and $1\times25$ to capture response patterns over different temporal scales. The smaller kernels represent local variations, whereas the larger kernels cover a wider temporal context. The resulting CNN features are concatenated with the extracted thermal features to form a 48-dimensional fused feature vector. This vector is then supplied to an ExtraTrees classifier to obtain the final material prediction. Fusion therefore occurs at the feature level, allowing the classifier to use both representations jointly when constructing its decision boundaries. This architecture combines explicitly extracted thermal information with patterns learned directly from the temperature sequence, providing complementary inputs for distinguishing materials with similar transient responses. Classification was evaluated using shuffled five-fold cross-validation with ten random seeds (2021--2030), with identical unstratified splits across observation windows and feature sets and all epoch selection, preprocessing, and feature selection confined to the training data. Table~\ref{tab:classification_time} reports the mean and sample standard deviation of the 50 test-fold accuracies.

Table~\ref{tab:classification_time} compares the three feature sets at observation times from $3$ to $15\ \text{s}$. At $3\ \text{s}$, fusion achieves $65.42\%$ accuracy, slightly exceeding CNN features ($65.03\%$); at $5\ \text{s}$, thermal features perform best ($87.86\%$ versus $86.14\%$ for fusion). Fusion yields the highest mean accuracy at all evaluated windows from $8$ to $15\ \text{s}$. Its peak of $92.69\pm3.59\%$ at $10\ \text{s}$ exceeds thermal and CNN features by $1.13$ and $5.91$ percentage points, respectively. Extending observation to $12$ and $15\ \text{s}$ gives $92.53\%$ and $92.61\%$, with no further improvement in mean accuracy. Thus, $10\ \text{s}$ offers a practical balance between classification accuracy and observation time.

\begin{figure}[!t]
    \centering
    \includegraphics[width=\columnwidth]{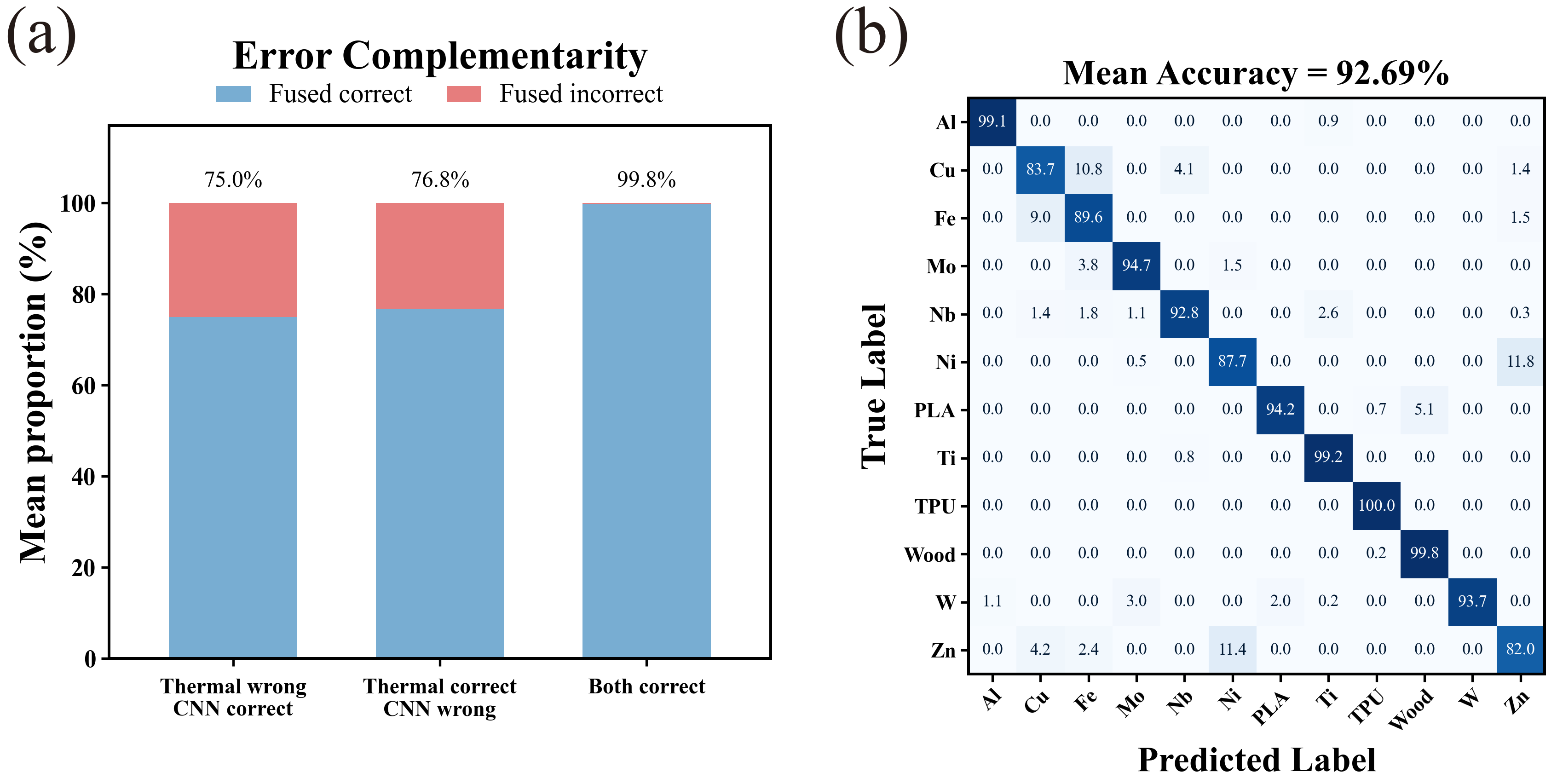}
    \caption{Classification performance at a 10 s observation window. (a) Mean conditional accuracy of the fused classifier, grouped by the correctness of the thermal-feature and CNN-feature classifiers. Blue and red segments indicate correct and incorrect fused predictions, respectively. (b) Confusion matrix of the fused classifier for 12 materials.}
    \label{Classification Results}
\end{figure}

Fig.~\ref{Classification Results}(a) examines the complementarity of the thermal-feature and CNN-feature models by grouping test predictions according to their correctness at the $10\ \text{s}$ observation window. When the thermal-feature model was incorrect and the CNN-feature model was correct, fusion achieved a mean recovery rate of $75.0\%$; conversely, when only the thermal-feature model was correct, the recovery rate was $76.8\%$. These results indicate that fusion benefits from both feature pathways, with a slightly higher recovery rate in the latter group. For predictions correctly classified by both models, fusion retained $99.8\%$ correct predictions on average, largely preserving their shared successes. Fig.~\ref{Classification Results}(b) shows the fused classifier's confusion matrix for 12 materials, with a mean test accuracy of $92.69\%$ and predictions concentrated along the main diagonal. TPU achieves $100.0\%$ recall, while Al, Ti, wood, PLA, W, and Nb attain $99.1\%$, $99.2\%$, $99.8\%$, $94.2\%$, $93.7\%$, and $92.8\%$, respectively. Prominent confusions occur between Cu and Fe and between Ni and Zn. Errors also include PLA predicted as wood ($5.1\%$) or TPU ($0.7\%$). Thus, the fused representation effectively distinguishes most tested materials, although confusion remains among certain metals and some nonmetallic classes.

\begin{table}[t]
\centering
\caption{Classification accuracy at different observation times}
\label{tab:classification_time}
\footnotesize
\setlength{\tabcolsep}{2pt}
\renewcommand{\arraystretch}{1.15}
\begin{tabular*}{\columnwidth}{@{}l@{\extracolsep{\fill}}ccc@{}}
\toprule
\textbf{Time} & \shortstack{\textbf{Thermal features}\\\textbf{Accuracy (\%)}} & \shortstack{\textbf{CNN features}\\\textbf{Accuracy (\%)}} & \shortstack{\textbf{Fused features}\\\textbf{Accuracy (\%)}} \\
\midrule
3 s & $58.00\pm5.86$ & $65.03\pm5.64$ & $65.42\pm5.38$ \\
5 s & $87.86\pm4.32$ & $81.89\pm4.51$ & $86.14\pm3.75$ \\
8 s & $90.36\pm3.78$ & $86.25\pm3.68$ & $91.28\pm3.09$ \\
10 s & $91.56\pm3.33$ & $86.78\pm3.95$ & $92.69\pm3.59$ \\
12 s & $90.97\pm3.50$ & $88.00\pm3.61$ & $92.53\pm3.49$ \\
15 s & $91.06\pm3.42$ & $89.33\pm3.22$ & $92.61\pm3.25$ \\

\bottomrule
\end{tabular*}
\end{table}

\subsection{Illustrative Gripper-Mounted Measurements}

\begin{figure}[htbp]
	\centering
	\resizebox{0.90\columnwidth}{!}{%
\begin{overpic}[width=\columnwidth,abs,unit=0.01\columnwidth]{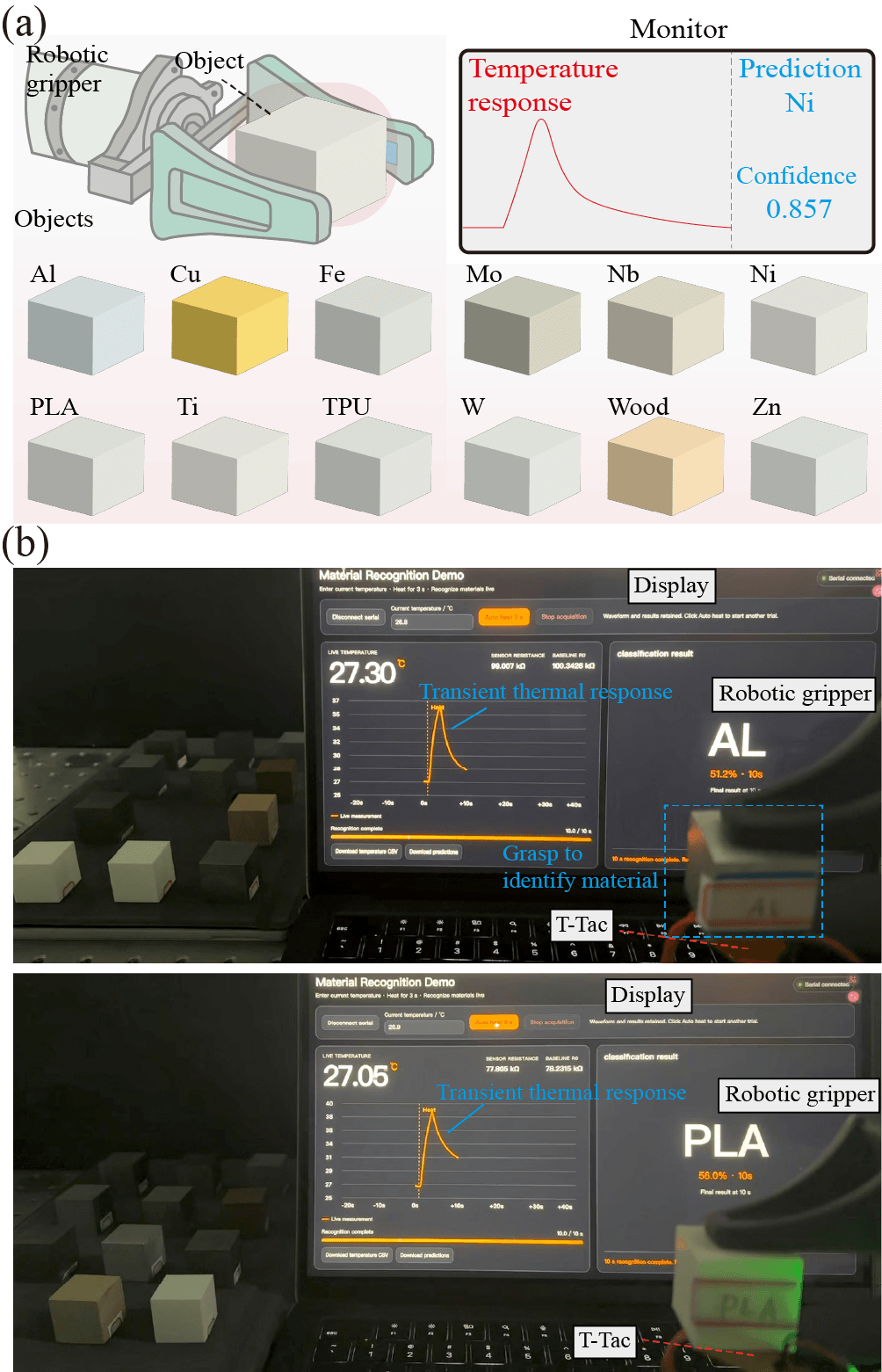}
        \put(62.4502,47.8088){\makebox(0,0)[bl]{\begingroup\setlength{\fboxsep}{0pt}\colorbox{white}{\parbox[c][2.7888\unitlength][c]{10.5578\unitlength}{\centering\fontsize{7.70}{8.20}\selectfont T-Tac}}\endgroup}}
        \put(63.1474,14.7410){\makebox(0,0)[bl]{\begingroup\setlength{\fboxsep}{0pt}\colorbox{white}{\parbox[c][2.7888\unitlength][c]{10.4582\unitlength}{\centering\fontsize{7.70}{8.20}\selectfont T-Tac}}\endgroup}}
    \end{overpic}%
}
    \caption{Gripper-mounted thermal sensing examples. (a) Schematic of the measurement and classification interface. (b) Photographs of gripper-mounted measurements with the acquired response and displayed model output.}
	\label{Validation on Material Identification}
\end{figure}

A host computer interface was developed to display the transient thermal response and classification output during gripper-mounted measurements. As shown in Fig.~\ref{Validation on Material Identification}(a), the system first performed alignment according to the current temperature. The gripper then grasped an object, and T-Tac measured the response to the heating pulse. A $10\ \text{s}$ segment starting from thermal excitation was supplied to the classification model, and the predicted material was displayed on the interface. Fig.~\ref{Validation on Material Identification}(b) shows examples of this measurement and display process. These examples illustrate the integration of T-Tac and its acquisition chain into a robotic gripper.

\begin{table*}[!t]
\centering
\caption{Representative thermal and multimodal tactile systems for material identification}
\label{tab:thermal_material_recognition}
\small
\setlength{\tabcolsep}{3.5pt}     %
\renewcommand{\arraystretch}{1.25} %
\begin{tabularx}{\textwidth}{
>{\raggedright\arraybackslash}p{1.2cm}   %
>{\raggedright\arraybackslash}p{4.9cm}   %
>{\raggedright\arraybackslash}p{4.9cm}   %
>{\raggedright\arraybackslash}p{2.2cm}   %
>{\raggedright\arraybackslash}p{1.8cm}   %
>{\raggedright\arraybackslash}X           %
}
\hline\hline
\textbf{Work} &
\textbf{Thermal mechanism} &
\textbf{Sensor structure} &
\textbf{Reported timing} &
\textbf{Reported result} &
\textbf{Mechanical form} \\
\hline
\cite{engelFlexibleMultimodalTactile2006} &
Active heating &
Au heater and Ni RTD on PI substrate &
NA &
\shortstack[l]{$>90\%$\\blind test} &
Flexible film \\

\cite{eguiluzMultimodalApproachContinuous2016} &
Heated BioTac with thermal and vibration sensing &
BioTac fingertip with thermal and vibration channels &
\shortstack[l]{0.28 s mean\\first recognition} &
\shortstack[l]{1.02\% error\\recursive\\protocol} &
\shortstack[l]{Rigid core,\\soft skin} \\

\cite{kerrMaterialRecognitionUsing2018} &
Preheated BioTac with thermal and vibration sensing &
BioTac fingertip with thermal and vibration channels &
\shortstack[l]{20 s pressing\\plus sliding} &
86.19\% &
\shortstack[l]{Rigid core,\\soft skin} \\

\cite{gomezeguiluzMultimodalMaterialIdentification2018} &
Heated BioTac with thermal and vibration sensing &
BioTac fingertip with thermal and vibration channels &
\shortstack[l]{0.28 s mean\\first recognition} &
98.6\% &
\shortstack[l]{Rigid core,\\soft skin} \\

\cite{bokTextureIdentificationObjects2021} &
Active heating (471 mW DC) &
Copper heater, gold RTD, and PDMS encapsulation &
\shortstack[l]{15 s static\\+ 10 s sliding} &
\shortstack[l]{98\%\\50 trials} &
Flexible sensing layer \\

\textbf{T-Tac} &
Active pulsed heating (single-sided TPS) &
PI heater and ionic film temperature sensor on FPCB substrate &
\shortstack[l]{3 s pulse\\10 s observation} &
92.69\% &
Film assembly \\
\hline\hline
\end{tabularx}
\par\smallskip
\parbox{\textwidth}{\footnotesize Reported times follow the original protocols and differ in exploratory actions and decision criteria. Accuracy and timing are not directly comparable across material sets.}
\end{table*}

\section{Comparison with Related Works}
\label{Comparison with Related Works}
Table~\ref{tab:thermal_material_recognition} compares representative thermal tactile systems in terms of thermal excitation, sensor structure, measurement protocol, and recognition performance. Flexible heater/resistance temperature detector (RTD) devices integrate heat generation and temperature sensing on a thin substrate \cite{engelFlexibleMultimodalTactile2006}. BioTac-based approaches combine thermal and vibration sensing within a fingertip comprising a rigid core and a soft skin \cite{eguiluzMultimodalApproachContinuous2016,kerrMaterialRecognitionUsing2018,gomezeguiluzMultimodalMaterialIdentification2018}. Other designs integrate a heater and RTD with a flexible sensing layer \cite{bokTextureIdentificationObjects2021}.

T-Tac combines a PI heater with an ionic conductive film temperature sensor in a film-based assembly. In the material-classification experiments, a $3\ \text{s}$ heating pulse excites the contact response, and changes in film resistance record the resulting heating and cooling transient. The classifier combines extracted thermal features with CNN time-series representations obtained from the same measured response. Table~\ref{tab:classification_time} reports a mean test-fold accuracy of $92.69\%$ for the 12 tested materials with a $10\ \text{s}$ observation window. The pulse duration specifies the thermal excitation, whereas the observation window defines the signal interval used for classification.

The studies in Table~\ref{tab:thermal_material_recognition} use different material sets and contact protocols, including pressing and sliding, and their reported times follow different decision criteria. Their results therefore characterize performance under their respective measurement conditions. The gripper-mounted examples illustrate how the film assembly and acquisition chain can be incorporated into robotic contact measurements. In this implementation, the sensing element is attached to the gripper surface to add thermal material sensing to the end effector.

\section{Conclusion}
\label{Conclusion}
In this article, we developed T-Tac, a soft active thermal tactile sensor that integrates a PI film heater with an ionic conductive film to realize single-sided transient thermal measurement in a compact and flexible form. By applying pulse heating and recording the transient thermal response, T-Tac provides active thermal excitation and sensitive temperature detection. The classification framework concatenates extracted thermal descriptors with CNN time-series features and uses an ExtraTrees classifier for material prediction. Experiments on the 12 tested materials achieved a mean test-fold accuracy of $92.69\%$ with a $10\ \text{s}$ observation window. Gripper-mounted measurements further illustrated the integration of the sensor and acquisition chain into a robotic end effector. These results support active thermal material classification under the reported measurement conditions. Future work will focus on investigating material recognition across objects with diverse volumes and surface conditions, pursuing greater generalizability in real-world scenarios, and extending the range of materials studied.

\ifCLASSOPTIONcaptionsoff
  \newpage
\fi

\bibliographystyle{IEEEtran}
\bibliography{references} 

\begin{IEEEbiography}[{\includegraphics[width=1in,height=1.25in,clip,keepaspectratio]{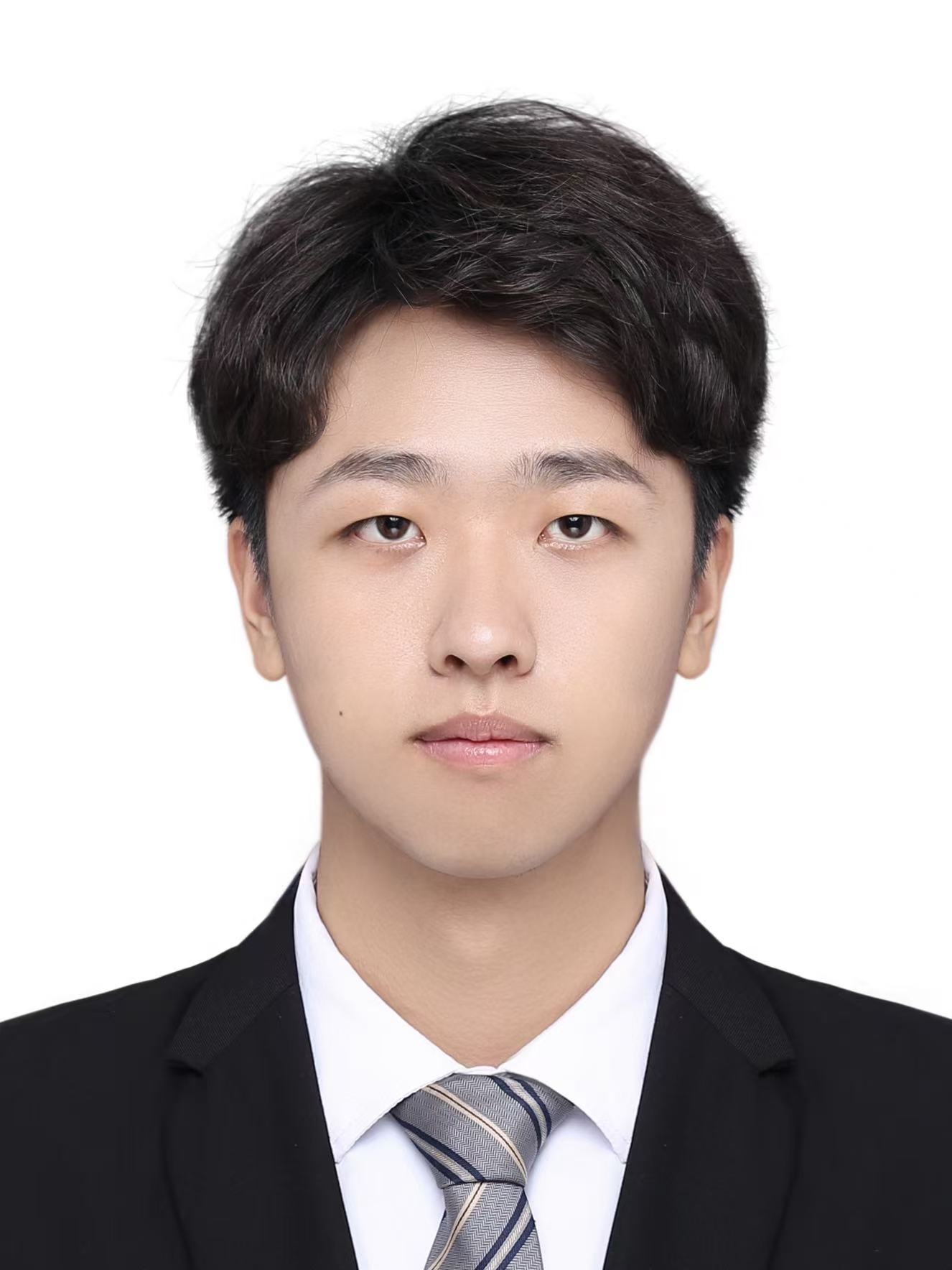}}]{Jiankun Li}
(Student Member, IEEE) received the bachelor's degree in automation engineering from Nanjing University of Aeronautics and Astronautics, Nanjing, China, in 2023, and the master's degree in Electronic Engineering from the University of Hong Kong, Hong Kong SAR, in 2024. He is currently pursuing a Ph.D. degree with the Department of Mechanical Engineering, City University of Hong Kong, Hong Kong SAR. His current research interests include tactile sensors and artificial intelligence.
\end{IEEEbiography}

\begin{IEEEbiography}[{\includegraphics[width=1in,height=1.25in,clip,keepaspectratio]{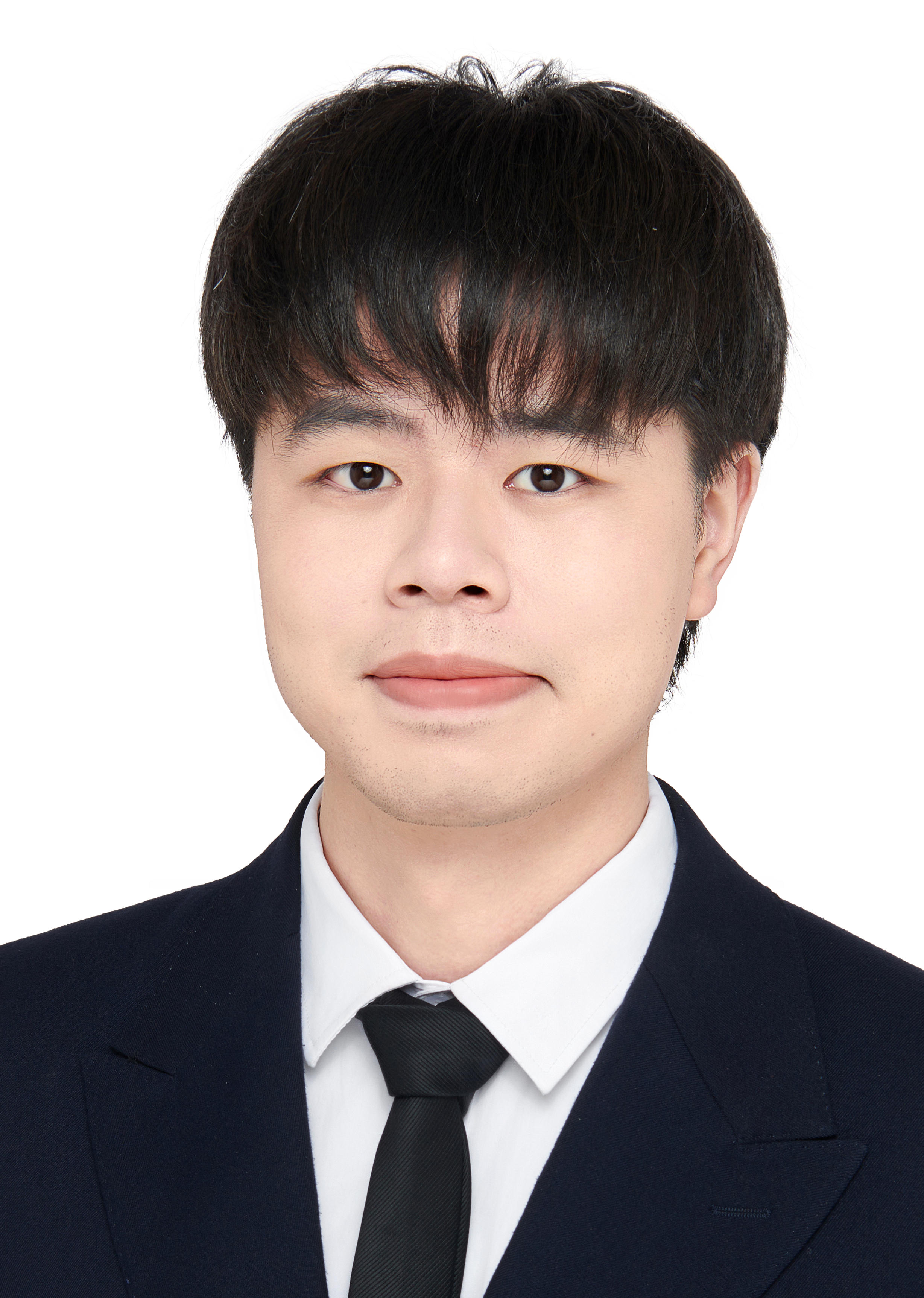}}]{Yu Feng}
(Member, IEEE) received the B.Eng. degree in Measurement and Automation from Harbin Engineering University, Harbin, China, in 2020, and the Ph.D. degree in Mechanical Engineering from City University of Hong Kong, Hong Kong SAR, in 2024. He subsequently worked as a postdoctoral fellow at City University of Hong Kong from 2024 to 2026. He is currently a research fellow with the Department of Mechanical Engineering, National University of Singapore, Singapore. His research interests include robotic perception, tactile sensors, and robot manipulation.
\end{IEEEbiography}

\begin{IEEEbiography}[{\includegraphics[width=1in,height=1.25in,clip,keepaspectratio]{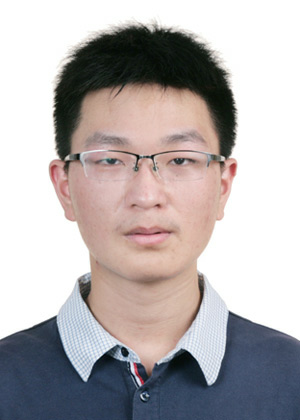}}]{Yehui Liu}
(Student Member, IEEE) received the B.S. degree in automation from Northeastern University at Qinhuangdao, Qinhuangdao, China, in 2024. He is currently pursuing the Ph.D. degree with the City University of Hong Kong, China. His current research interests include machine learning and sensor signal processing.
\end{IEEEbiography}

\begin{IEEEbiography}[{\includegraphics[width=1in,height=1.25in,clip,keepaspectratio]{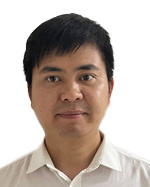}}]
{Meng Chen}
received the B.S. degree in software engineering from Shenzhen University, Shenzhen, China, in 2013, and the Ph.D. degree in mechanical engineering from the City University of Hong Kong, Hong Kong, in 2021. He is currently a Postdoctoral Fellow with the City University of Hong Kong. His current research interests include sensor networks, motion analysis, and artificial intelligence.
\end{IEEEbiography}

\begin{IEEEbiography}[{\includegraphics[width=1in,height=1.25in,clip,keepaspectratio]{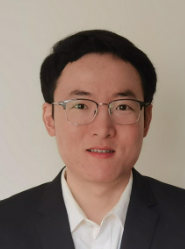}}]
{Rui Luo}
earned his BEng from Tsinghua University, and his PhD degree from Cornell University. His research interests include statistical signal processing and complex networks. Currently, Rui Luo is an Assistant Professor in the Department of Systems Engineering at City University of Hong Kong. His current research interests include network analysis, machine learning, and intelligent systems.
\end{IEEEbiography}

\begin{IEEEbiography}[{\includegraphics[width=1in,height=1.25in,clip,keepaspectratio]{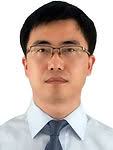}}]{Guanglie Zhang}
(Senior Member, IEEE) received the B.E. and M.E. degrees in electronic engineering and the Ph.D. degree in control science and engineering from Xi'an Jiaotong University, Xi'an, China, in 1997, 2000, and 2003, respectively. He is currently a Research Associate Professor with the Department of Mechanical Engineering, City University of Hong Kong, Hong Kong SAR. His research interests cover areas of nanotechnology, motion recognition, and complex-sensor-network fusion.
\end{IEEEbiography}

\begin{IEEEbiography}[{\includegraphics[width=1in,height=1.25in,clip,keepaspectratio]{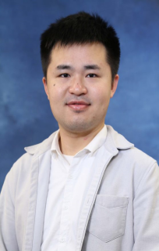}}]{Jianshu Zhou} (Member, IEEE) received the Ph.D. degree in Mechanical Engineering from The University of Hong Kong in 2020. He is currently an Assistant Professor with the Department of Mechanical Engineering at National University of Singapore. Prior to this, he was a Postdoctoral Scholar with the Mechanical Systems Control Laboratory, Department of Mechanical Engineering, University of California, Berkeley. He also served as a Research Assistant Professor with The Chinese University of Hong Kong. His research interests include robotics and mechatronics, robotic grasping and manipulation, embodied intelligence and physical AI, and multimodal sensing, data acquisition, and physical interaction.
\end{IEEEbiography}

\begin{IEEEbiography}[{\includegraphics[width=1in,height=1.25in,clip,keepaspectratio]{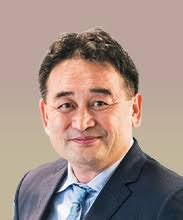}}]{Wen Jung Li}
(Fellow, IEEE) received the B.S. and M.S. degrees in aerospace engineering from the University of Southern California, Los Angeles, CA, USA, in 1987 and 1989, respectively, and the Ph.D. degree in aerospace engineering from the University of California at Los Angeles in 1997. He worked in the Department of Mechanical and Automation Engineering, the Chinese University of Hong Kong (CUHK), from 1997 to 2011. Prior to CUHK, he held research and development positions with the NASA/Caltech Jet Propulsion Laboratory, Pasadena, CA, USA; The Aerospace Corporation, El Segundo, CA, USA; and Silicon Microstructures Inc., Fremont, CA, USA. He is honored as an International Fellow of the Canadian Academy of Engineering (CAE), a Member of the US National Academy of Artificial Intelligence (NAAI), a Fellow of European Academy of Engineering (EAE), an IEEE Fellow, an ASME Fellow, an Asia-Pacific AI Association (AAIA) Fellow, and a 100 Talents Program awardee of the Chinese Academy of Sciences. He served as the President of the IEEE Nanotechnology Council in 2016 and 2017. He currently holds a Chair Professorship of the Department of Mechanical Engineering and serves as the Vice-President (Talent and International Strategy) at City University of Hong Kong, Hong Kong. His research interests encompass AI-enabled sensors, multimodal tactile sensing, and intelligent cyber-physical sensors.
\end{IEEEbiography}

\end{document}